 \documentclass[smallabstract,smallcaptions]{dccpaper}

\usepackage{epsfig}
\usepackage{citesort}
\usepackage{amsmath}
\usepackage{amssymb}
\usepackage{color}
\usepackage{url}
\usepackage{subfigure}
\newlength{\figurewidth}
\newlength{\smallfigurewidth}
\usepackage{multirow}
\usepackage{caption}
\usepackage{soul}
\usepackage{nameref}

\newcommand{\tabincell}[2]{\begin{tabular}{@{}#1@{}}#2\end{tabular}}

\begin{document}

\title
{\large
\textbf{High-fidelity 3D Model Compression based on Key Spheres}
}


\author{%
Yuanzhan Li, Yuqi Liu, Yujie Lu, Siyu Zhang, Shen Cai$^{\ast}$, and Yanting Zhang  \\
{\small\begin{minipage}{\linewidth}\begin{center}
\begin{tabular}{c}
Visual and Geometric Perception Lab, Donghua University,
Shanghai, China \\
$^{\ast}$Correspondence: \url{hammer_cai@163.com} 
\end{tabular}
\end{center}\end{minipage}}
}

\maketitle
\thispagestyle{empty}

\renewcommand{\thefootnote}{}

\footnote{\textbf{Acknowledgment}\quad This work is supported by Shanghai Sailing Program (21YF1401300), Natural Science Foundation of Shanghai (Grant No. 21ZR1401200), and the Foundation of Key Laboratory of Artificial Intelligence, Ministry of Education, P.R. China (AI2020003).}
\begin{abstract}
In recent years, neural signed distance function (SDF) has become one of the most effective representation methods for 3D models. 
By learning continuous SDFs in 3D space, neural networks can predict the distance from a given query space point to its closest object surface, whose positive and negative signs denote inside and outside of the object, respectively.
Training a specific network for each 3D model, which individually embeds its shape, can realize compressed representation of objects by storing fewer network (and possibly latent) parameters. 
Consequently, reconstruction through network inference and surface recovery can be achieved. 
In this paper, we propose an SDF prediction network using explicit key spheres as input. 
Key spheres are extracted from the internal space of objects, whose centers either have relatively larger SDF values (sphere radii), or are located at essential positions. 
By inputting the spatial information of multiple spheres which imply different local shapes, the proposed method can significantly improve the reconstruction accuracy with a negligible storage cost. 
Compared to previous works, our method achieves the high-fidelity and high-compression 3D object coding and reconstruction. 
Experiments conducted on three datasets verify the superior performance of our method.
\end{abstract}

\Section{1. Introduction}
\vspace{-3mm}
3D shape representation is one of the core issues in computer graphics and computer vision. Conventional representations are generally explicit, such as surface points, mesh and voxels, etc. Due to their discreteness, it is difficult to directly compress data based on these representations. 
For surface points, in order to reduce the amount of data stored or processed without sacrificing the performance in vision or graphics tasks, a variety of sampling methods have been studied, such as the traditional farthest point sampling (FPS)~\cite{FPS}, the semantic segmentation based on randomly sampled point clouds~\cite{RandLANet} and the learning based sampling~\cite{S_Net}.
However, these methods are explicit and perform inconsistently in different tasks. 
For mesh, there are also some works studying mesh coarsing and simplification, such as the traditional quadric edge collapse decimation (QECD)~\cite{meshlab} and a deep learning based simplification~\cite{meshcnn}.
These representations are still explicit and at the risk of losing details of objects. 
For voxels, octree representation is a feasible approach to data compression, which has been adopted in both traditional object representation~\cite{Octree} and deep learning based object classification~\cite{OCNN}. 
However, for complex objects, there is still a trade-off between fine details and data compression in voxel octree representation.

Recently, the development of deep learning has made it possible to encode 3D models through network parameters (and possibly latent parameters). 
This category of methods is called implicit neural representation of geometric shapes. 
For example, DeepSDF~\cite{deepsdf} early reconstructs one class of 3D objects using a large multi-layer perception (MLP) network to learn continuous SDFs.
Along this line, FFN~\cite{FFN} improves the fitting capability of MLP by mapping Fourier features and learning high-frequency functions in low-dimensional domains. 
Similarly, SIREN~\cite{SIREN} enhances the representation capability of the network by modifying the activation function in MLP. 

\vspace{1mm}
With the increase in the number and diversity of samples within a class, the shape representation network for one class becomes more difficult to train. 
To overcome this problem, NI~\cite{NI} firstly presents an overfitting MLP network which is individually trained for an object to embed its global shape. 
This weight-encoded
implicit neural representation actually accomplishes a lossy compression for each 3D model.
However, the default network with 7553 parameters may suffer a large reconstruction error for a complex object.
NGLOD~\cite{nglod} proposes to fit local shapes of an object by inputting pre-trained latent vectors of octree vertices into the network.
Although this local-fitting network promotes reconstruction accuracy, the storage capacity is greatly increased as the latent feature vectors of large number of grid points needs to be stored, in addition to the network parameters.

\vspace{1mm}
It is worth mentioning that the recent popular NeRF works~\cite{NeRF}~\cite{Giraffe} solve a problem different from the above implicit neural methods.
Although NeRFs encode the 3D scenes including shapes, texture and illumination also using a MLP network, their aim is to reconstruct objects from a set of 2D images. Therefore, NeRFs are nearly irrelevant to the compression of a known 3D model which is the focus of this article.

\begin{figure*}[t]
	\centering
	\includegraphics[width=1\textwidth]{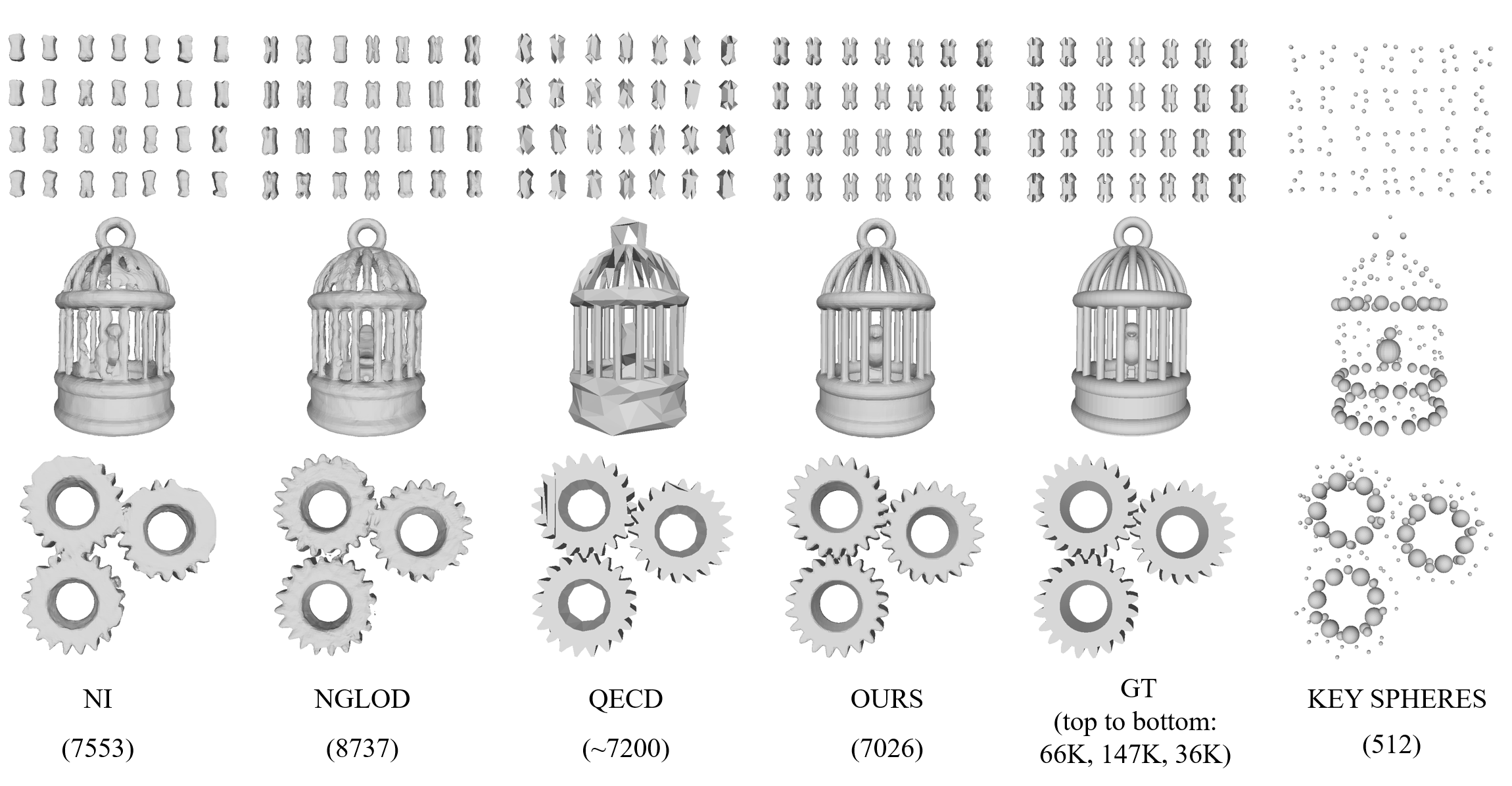}
	\vspace{-9mm}
	\caption{Comparison of three implicit neural reconstruction methods (NI, NGLOD and OURS), a mesh simplification method (QECD) and ground truth (GT). 
	The number of storage parameters is shown in parentheses.
	An MLP network with 8 hidden layers and $32$ nodes in each layer is used in NI.
	The $32$-dimensional latent vectors of $125$ grid points and $4737$ network parameters are stored in NGLOD.
    In QECD, the number of vertices and facets is manually controlled at about $2400$, which represents about $7200$ parameters.
    Three ground truth models with about 22K, 49K, 12K vertices and facets respectively, are shown in the fifth column from left to right. 
	The 128 extracted spheres with 512 parameters, which are used as the input of our network, are displayed in the last column.
	Except for our method with the fewest parameters ($7026$), all other methods show obvious errors. }
	\label{fig:four_method}
	\vspace{-5mm}
\end{figure*}

\vspace{1mm}
In this paper, we propose a signed distance function (SDF) prediction network combining with explicit key spheres input.
Key spheres presented in one of our previous works~\cite{SN} are extracted from the inside of the object, whose centers either have relatively larger SDF values (i.e., sphere radii), or are located at essential positions. 
Although a sphere node graph of each model is initially constructed as a concise representation for the object classification task, we discover that key spheres can play an important role in fitting SDFs of 3D shapes.
By explicitly inputting spatial information of multiple key spheres, the proposed method using global shape fitting can significantly reduce the reconstruction error, even compared with the state-of-the-art local fitting method~\cite{nglod}.
Our method can be considered as a hybrid neural representation to some extent, which utilizes explicit information at a small additional storage cost, but greatly improves the reconstruction accuracy of global implicit representation.
Fig.~\ref{fig:four_method} shows the reconstruction results of four methods and the ground truth models, which belong to the Thingi10K~\cite{Thingi10k} dataset.
The results (shown in the first two columns from left to right) of two implicit neural methods NI and NGLOD tend to lack details.  
The result (shown in the third column) of the traditional mesh simplification method QECD coarsens the three models obviously.
Our reconstruction results given in the fourth column show no noticeable errors. 
In addition, compared with other methods, the number of storage parameters we used is the fewest. 
The $128$ extracted key spheres, which visually imply the rough model shape, are also displayed in the last column. 
From the viewpoint of shape coding and valid recovery, only the proposed method achieves the high-fidelity 3D model compression.

Our contributions are summarized as follows:
\vspace{-2mm}
\begin{itemize}
  \item We propose a signed distance function (SDF) prediction network based on explicit key spheres input, which naturally contains the rough shape information.
  \item We reduce neural network complexity and SDF fitting difficulty at a neglected additional storage cost (centers and radii of multiple key spheres).
  \item We experimentally verify that the reconstruction accuracy of the global fitting way (our approach) can surpass the local fitting strategy (NGLOD).
\end{itemize}

\vspace{-3mm}
\Section{2. Key Spheres based Model Compression}
\vspace{-3mm}
The purpose of our method is to utilize explicit key spheres containing rough shape information to directly realize high-precision global shape reconstruction. 
Specifically, we individually train a network with the same configuration for each object. 
Ideally, the trained network can overfit the SDF of an object, and the parameters of the SDF-fitting network and multiple key spheres are stored.
As a result, through the network inference which outputs the SDF value of every query point, the surface of the 3D model can be recovered by taking advantage of a post-processing algorithm.

This section describes the proposed method in detail.
Sec.~\nameref{3.1} introduces the background of SDF and key spheres. Sec.~\nameref{3.2} explains the origin of our core idea, and the advantages of implicit neural representation based on key spheres. Sec.~\nameref{3.3} shows the proposed network architecture. Sec.~\nameref{3.4} provides other details of the overview of the network.

\SubSection{2.1 Signed Distance Functions and Key Spheres}\label{3.1}
\vspace{-1mm}
Signed distance function (SDF) is one kind of implicit expression of 3D models, which has been often used in 3D object classification and reconstruction.
We assume that a normalized 3D object $\mathcal{V}$ is inside a space denoted by $\mathcal{H}\!=\![-1,1]^3$. The value of the function $SDF(\mathbf{x})$ is defined as the minimum distance from a query point $\mathbf{x}$ in $\mathcal{H}$ to any point $\mathbf{p}$ on the object surface $\mathcal{M}$,
\begin{equation}
\vspace{-1mm}
\mathop{SDF(\mathbf{x})}\limits_{\mathbf{x} \in \mathcal{H}} = sign(\mathbf{x}) \min\limits_{\mathbf{p} \in \mathcal{M}} \|\mathbf{x}-\mathbf{p}\|
\label{equ.1}
\vspace{-1mm}
\end{equation}
where the signed function $sign(\mathbf{x})$ is equal to $1$ if $\mathbf{x}\!\in\!\mathcal{V}$, otherwise equal to $0$. Since the surface $\mathcal{M}\!\in\!\mathcal{V}$, if $\mathbf{x}$ is a surface point, $sign(\mathbf{x})\!=\!1$ and $\|\mathbf{x}-\mathbf{p}\|\!=\!0$. Considering the neural network as a function $Net(\mathbf{x})$, the goal of fitting is that $Net(\mathbf{x})$ is approximate to $SDF(\mathbf{x})$ as much as possible, which is similar to the previous methods ~\cite{NI}~\cite{nglod}.

Key spheres is a concise expression method \cite{InSphereNet}~\cite{SN} for 3D models, which extracts a certain number of spheres based on SDF. 
For each sphere, its center can be located at any point $\mathbf{x}$ in space $\mathcal{H}$, and its radius is the SDF value of the center, i.e., $SDF(\mathbf{x})$.
Thus a surface point also can be considered as a degenerate sphere with zero radius.
In principal, a number of spheres can be selected randomly, uniformly (like grid vertices), or according to some specific rules.
Key spheres used in this paper as explicit inputs to the network come from one of our previous works~\cite{SN}, which takes into account sphere radii and mutual distances to select spheres.
For most objects, the rough shape and structure can be seen only with a few to dozens of key spheres. 
The three examples of 128 key spheres have been shown in Fig.~\ref{fig:four_method}. More examples could be seen in the website~\cite{SNG_git}.

\SubSection{2.2 Implicit Neural Representation with Key Spheres}
\label{3.2}
\vspace{-1mm}
Implicit neural representation usually uses multi-layer perception (MLP) networks to fit global SDF~\cite{deepsdf}~\cite{NI} or local SDF~\cite{nglod} of an object. 
Universal approximation theorem~\cite{universal_approximator} has proved that feed forward neural network with a single hidden layer and a finite number of neurons can fit any complexity function with any accuracy. 
However, for complex 3D models or simple scenes with multiple objects (see Fig.~\ref{fig:four_method}), it is difficult for MLP to successfully fit their global SDFs.
Although the above-mentioned methods improve the local reconstruction quality by fitting SDF on object blocks, they all need to store a large number of latent vectors in advance, which may even exceed the storage consumption of the original mesh model.

\begin{figure}[t]
	\centering
	\includegraphics[width=0.75\textwidth]{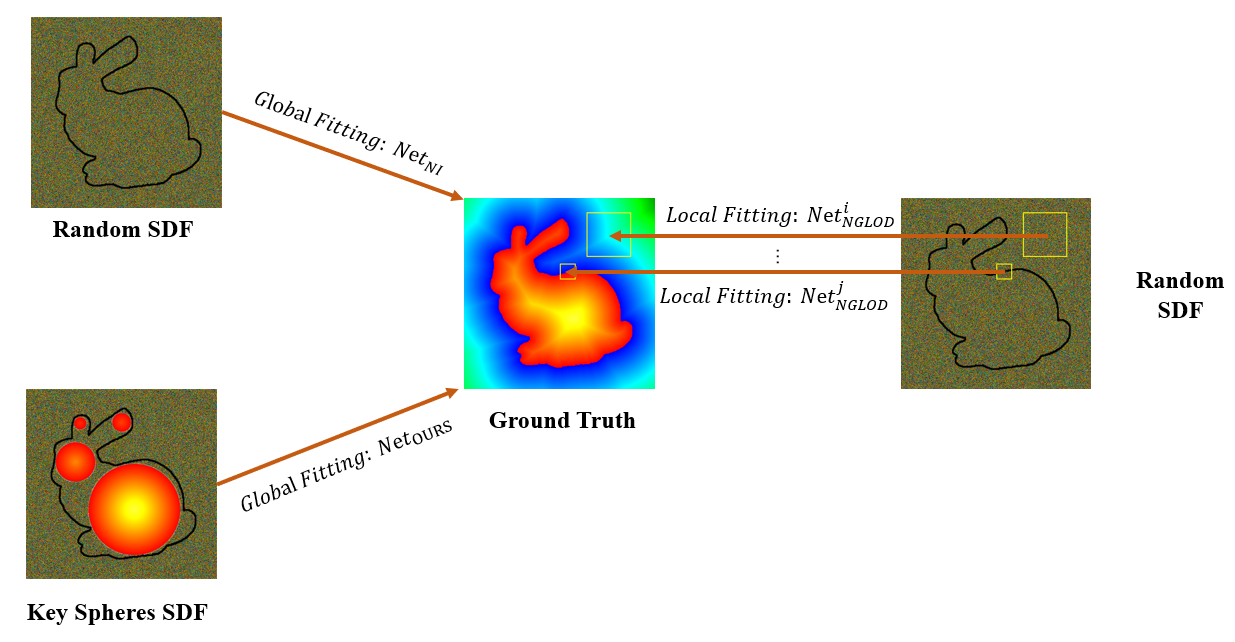}
	\vspace{-3mm}
	\caption[width=1\textwidth]{Fitting SDF comparison of three methods. Ground truth SDF is depicted in pseudo color in the middle. The constrained SDFs of NI, NGLOD and key spheres are shown on the right and on the left, respectively. The corresponding global or local fitting networks are denoted by $Net_{NI}$, $Net_{NGLOD}$ and $Net_{OURS}$, respectively.} 
	\vspace{-5mm}
	\label{fig:SDF_rabbit}
\end{figure}

To effectively improve the reconstruction performance and reduce the storage cost, we introduce an implicit neural representation with explicit key spheres inputs. 
We observe that the visualized key spheres occupy most of the object space. Thus employing the key spheres can greatly reduce the difficulty of fitting SDF.
In Fig.~\ref{fig:SDF_rabbit}, the SDF of a 2D Bunny model is drawn in the middle. 
Supposing four key spheres are extracted from the object, the SDF value of any point inside a sphere $\mathcal{S}_j$ is directly constrained by this key sphere.
Denote the sphere center by $\mathbf{c}_j$. 
There exists a constraint for the point inside the key spheres, which is expressed by
\vspace{-2mm}
\begin{equation}
    rad(\mathcal{S}_j) \geq SDF(\mathbf{x}) \geq rad(\mathcal{S}_j) - \|\mathbf{x}\!-\!\mathbf{c}_j\|, \quad \mathrm{if} \:\: \mathbf{x}\!\in\!\mathcal{S}_j
\label{equ.2}
\vspace{-2mm}
\end{equation}
where the radius $rad(\mathcal{S}_j)$ is equal to $SDF(\mathbf{c}_j)$. 

The lower bounds of SDF values of points inside the four spheres are drawn on the bottom left of Fig.~\ref{fig:SDF_rabbit}. 
SDF values of other points outside the spheres can be considered as unknown random numbers (drawn with a random color), similar to the output of neural network with random initial parameters. 
Thus, the idea of our method can be simply regarded as learning a global mapping from the bottom left SDF to the middle SDF through the proposed network $Net_{OURS}$. 
For comparison, the SDF fitting process of NI~\cite{NI} and NGLOD~\cite{nglod} are also drawn on the top left and on the right of Fig.~\ref{fig:SDF_rabbit}, respectively. 
NI directly lets the network $Net_{NI}$ learn the global mapping from the overall random SDF to the model SDF.
Therefore, it can be foreseen that NI is likely to fail for complex objects as the fitting ability of several MLP layers is pretty limited.
NGLOD adopts a local fitting strategy to alleviate this problem.  
The interpolation of latent vectors of grid vertices (e.g., $Net_{NGLOD}^{i}$ and $Net_{NGLOD}^{j}$) is adopted to fit the corresponding local SDF subject to different size blocks divided by the octree (e.g., the yellow squares).

Intuitively, the explicit key spheres input can be considered as a good upper and lower bound of SDF for most of object space in global optimization. 
Thus the proposed method can fit the global SDF of an object more easily than NI, at the cost of storing the information of multiple key spheres, which can be neglected compared to storing network parameters.
Compared with NGLOD, there are some advantages of the proposed method. Firstly, the storage demand is much smaller than NGLOD’s, i.e., the number of key spheres is much less than the number of octree vertices.
Secondly, adding the SDF value makes a point into a sphere, which contains rough shape information.
This is why the proposed global fitting network performs better than the existing local fitting networks under almost the same number of parameters.

\begin{figure}[t]
	\centering
	\includegraphics[width=0.95\textwidth]{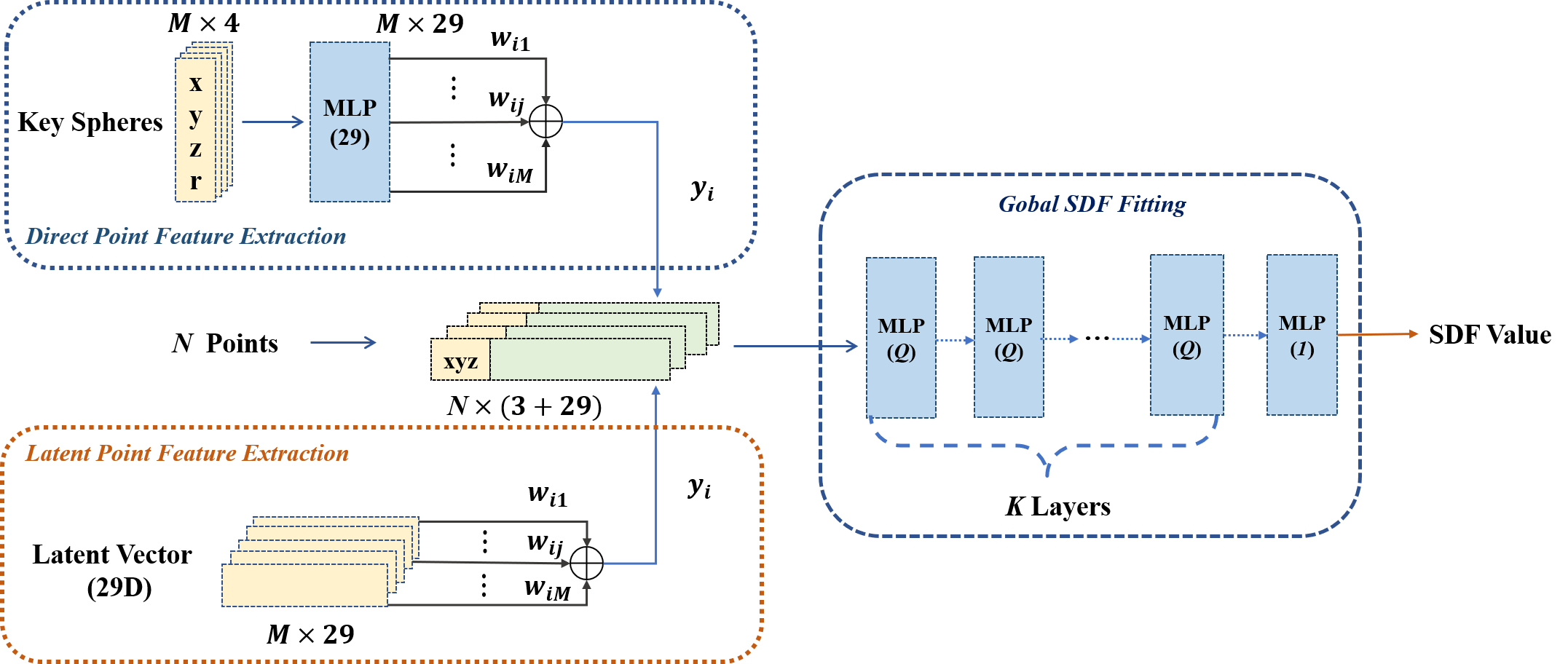}
	\vspace{-2mm}
	\caption{Proposed network architecture. To extract 29D point features, two branches are designed as direct and latent ways, respectively.
	\label{fig:net_architecture}}
	\vspace{-5mm}
\end{figure}

\vspace{-1mm}
\SubSection{2.3 Network Architecture}\label{3.3}
\vspace{-2mm}
The proposed network architecture is shown in Fig.~\ref{fig:net_architecture}. 
There are $N$ points with 3D coordinates (xyz) and $M$ key spheres with 4D vectors (xyzr).
The 4D vector of $j$-th key sphere refers to sphere center coordinates $\mathbf{c}_j$ and radius of $\mathcal{S}_j$.
The global fitting part shown in the middle of the right side is a $K$-layer MLP network with the same $Q$ nodes to fit SDF of each model.
In addition to the 3D coordinates for $i$-th query point, a 29-dimensional (29D) point feature vector $\mathbf{y}_i$ is concatenated as a part of input to MLP fitting layers.
In order to make a fair comparison with NI and NGLOD respectively, this 29D point feature can be extracted in direct and latent ways based on key spheres.
The direct point feature extraction (DPFE, see the upper branch of Fig.~\ref{fig:net_architecture}) only uses a single-layer MLP ($4\!*\!29$) to upgrade the 4D input of each key sphere to a 29D feature $\mathbf{z}_j$.
The 29D feature vectors of all key spheres are linearly weighted with the coefficient $w_{ij}$ to generate the point feature vector $\mathbf{y}_i$,
\vspace{-3mm}
\begin{equation}
     \mathbf{y}_i \!=\! \sum\limits_{j=1}^{M}(w_{ij} \!*\! \mathbf{z}_j), \quad \mathrm{with} \; w_{ij} \!=\! {d_{ij}} \,/\, {\sum\limits_{j=1}^{M}d_{ij}}, \quad \mathrm{where} \; d_{ij} \!=\! 2\!*\!rad(\mathcal{S}_j) \!+\! dis(\mathbf{x}_i, \mathbf{c}_j).
    \label{equ.5}
\end{equation}
\vspace{-5mm}

\noindent where the weight $w_{ij}$ is jointly determined by the distances from the query point $\mathbf{x}_i$ to the sphere centers and the radii of all $M$ key spheres,

In the DPFE branch, compared with NI, the proposed network simply adds one-layer MLP to extract point feature vectors. The total number of storage parameters of the direct method is shown in the following equation. 
For example, in the network used to reconstruct the object shown in Fig.~\ref{fig:four_method}, we set the number of key spheres $M\!=\!128$,  the number of hidden layers $K\!=\!6$, and the number of neurons in each layer $Q\!=\!32$. Thus the total number of storage parameters is $7026$.
\vspace{-3mm}
\begin{equation}
\underbrace{M*4}_\text{key spheres input}+ 
\underbrace{29*4}_\text{point feature layer} + \underbrace{Q*K+30}_\text{total offset}+\underbrace{Q*Q*(K-1)+(Q+1)*32}_\text{global fitting layers}
\end{equation}
\vspace{-5mm}

The latent point feature extraction (LPFE, see the lower branch in Fig.~\ref{fig:net_architecture}) 
is similar to the latent feature of grid points in NGLOD. The 29D sphere feature vector $z_j$ is obtained by training, which is stored in advance. 
These stored sphere feature vectors are linearly weighted with the coefficient $w_{ij}$ in inference to generate the point feature vector $\mathbf{y}_i$. 
In LPFE, the network parameters are slightly reduced as the $4*29$ dimension point feature layer is removed. But the storage parameters are obviously increased by $M*29$. 
The experimental performances of both branches are similar. 
We use the DPFE method by default because it is more intuitive and has a smaller number of parameters.
The LPFE method is only used in Sec.~\nameref{4.3} when the comparative experiments mainly between our method and NGLOD are conducted.

\SubSection{2.4 Other Network Configurations}\label{3.4}
\vspace{-2mm}
The workflow of the proposed method is basically the same as NI.
1M points from a normalized space are sampled, whose positions and SDF values are computed as a training set.
Key spheres are extracted by using the method~\cite{SN}.
The loss function in training is $L_1$ distance of SDF values. 
Surface reconstruction uses the traditional marching cube method~\cite{Marching_cubes} as a post-processing step.

\Section{3. Experiments}
\label{section:4}
\vspace{-3mm}
Various experiments are conducted with
the default global fitting network which has $6$ hidden layers and $32$ neurons in each layer.
Datasets used in this paper are Thingi10K~\cite{Thingi10k}, Thingi32 and ShapeNet150. 
Thingi10K contains 10,000 3D-printing models.
Thingi32 is composed of 32 simple shapes in Thingi10K. ShapeNet150 contains 150 shapes in the ShapeNet dataset~\cite{shapeNet}, including 50 cars, 50 airplanes, and 50 chairs. 
We use four metrics for evaluation. The first two metrics defined in NI~\cite{NI} are surface error and importance error. The latter two metrics used in NGLOD~\cite{nglod} are chamfer distance (CD) and generalized intersection over union (gIoU). 

\textbf{All the shown results can be reproduced by the code on GitHub~\cite{ModelCompression}. Meanwhile, the uploaded material also includes all the shown 3D models and the pre-trained network models.}

\begin{figure}[tbp]
\begin{tabular}{cc}
\begin{minipage}[t]{0.48\linewidth}
    \includegraphics[width = 1\linewidth]{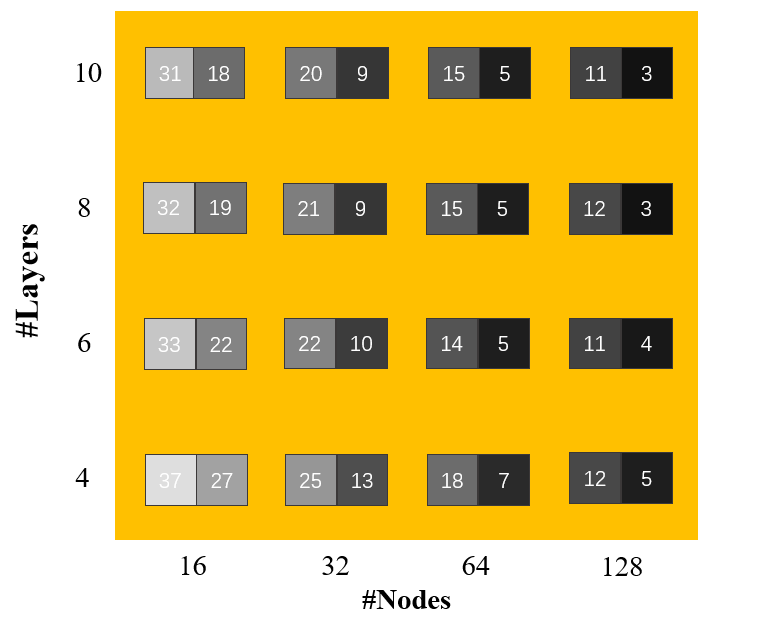}
    \vspace{-8mm}
    \caption{Surface error (times 10,000) of reconstructed models of NI (shown at the front) and our method (shown at back) on Thingi32 dataset under different numbers of layers and hidden nodes. 
	The color in the square becomes dark as the number decreases. 
	Apparently, each method gains lower error with the increased network size and our method outperforms NI under any configuration.
	}
	\label{fig:layer_hidden}
	\begin{center}
    \resizebox{1\textwidth}{!}{
    \begin{tabular}{|l||c|c|c|c|c|}
    \hline
    \multirow{2}{*}{Method}  & \multirow{2}{*}{\tabincell{c}{Storage\\(KB)}} & \multicolumn{2}{c|}{ShapeNet150} & \multicolumn{2}{c|}{Thingi32} \\
    \cline{3-6}
    & & gIoU & CD & gIoU & CD 
    \\
    \hline
    DeepSDF  & 7186 & 86.9 & 0.316 & 96.8 & 0.053         \\ \hline 
    FFN  & 2059 & 88.5 & 0.077 & 97.7 & 0.033   \\\hline
    SIREN  & 1033  &  78.4 & 0.381 &95.1 & 0.077 \\ \hline
    NI  &  30 &  82.2 & 0.5 & 96 & 0.092 \\ \hline\hline
    \tabincell{c}{NGLOD:\\(LOD1)}  & \tabincell{c}{96} & \tabincell{c}{84.6}  & \tabincell{c}{0.343}    & \tabincell{c}{96.8}  & \tabincell{c}{0.079}\\\hline
    \tabincell{c}{\:(LOD2)}  & \tabincell{c}{111} & \tabincell{c}{88.3}  & \tabincell{c}{0.198}    & \tabincell{c}{98.2}  & \tabincell{c}{0.041}\\\hline
    \tabincell{c}{\:(LOD3)}  & \tabincell{c}{163} & \tabincell{c}{90.4}  & \tabincell{c}{0.112}    & \tabincell{c}{99}  & \tabincell{c}{0.030}\\\hline
    \tabincell{c}{\:(LOD5)}  & \tabincell{c}{1356} & \tabincell{c}{91.7}  & \tabincell{c}{0.062}    & \tabincell{c}{\textbf{99.4}}  & \tabincell{c}{0.027}\\\hline\hline
    \tabincell{c}{OURS:\\(32)} &  30  & 93.8 & 0.188 & 98.6 & 0.028 \\ \hline
    \;\;(128) & 42  & 93.8 &  0.138  &  98.6 & 0.029 \\ \hline 
    \;\;(512) & 93 & 94.1 & 0.124 & 98.5 & 0.031 \\ \hline
    \tabincell{c}{\;\;(512*)}  & \tabincell{c}{158} & \tabincell{c}{\textbf{95.9}}  & \tabincell{c}{\textbf{0.060}}    & \tabincell{c}{99.2}  & \tabincell{c}{\textbf{0.025}}\\\hline
    \end{tabular}
    }
    \vspace{-2mm}
    \captionof{table}{Storage and reconstruction metrics of multiple methods on ShapeNet150 and Thingi32 datasets. For NGLOD and our method, the results with four different octree depths and sphere numbers are shown in $6$-th to $9$-th rows and $10$-th to $13$-th rows, respectively. `*' denotes that the network with 6 layers and 64 hidden nodes is adopted, instead of the default network configuration.}
    \label{tab:nglod} 
    \end{center}
\end{minipage}

\quad

\begin{minipage}[t]{0.48\linewidth}
    \includegraphics[width = 1\linewidth]{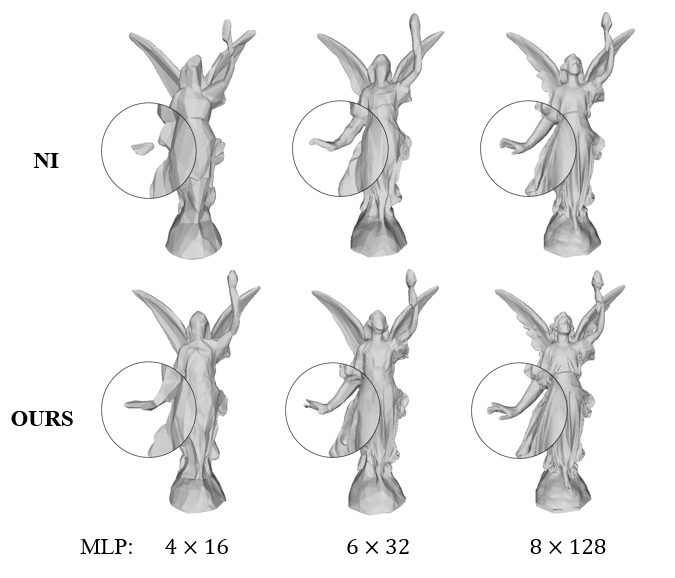}
    \vspace{-8mm}
    \caption{The reconstructed statues of NI and our method under three different global fitting network configurations, respectively. Numbers shown below the figures indicate layers and hidden nodes of the adopted MLP. The original mesh model has about $224$K parameters.}
	\label{fig:Statue}
	\centering
	\vspace{2mm}
	\subfigure[Surface error]{
    \begin{minipage}[b]{0.47\textwidth}
    \includegraphics[width=1\textwidth]{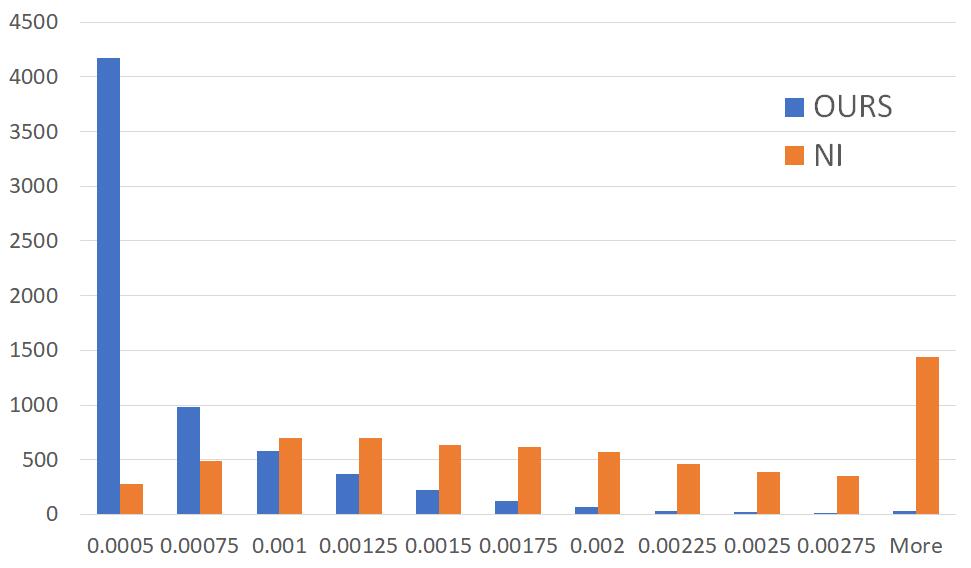}
    \end{minipage}}
    \subfigure[Importance error]{
    \begin{minipage}[b]{0.47\textwidth}
    \includegraphics[width=1\textwidth]{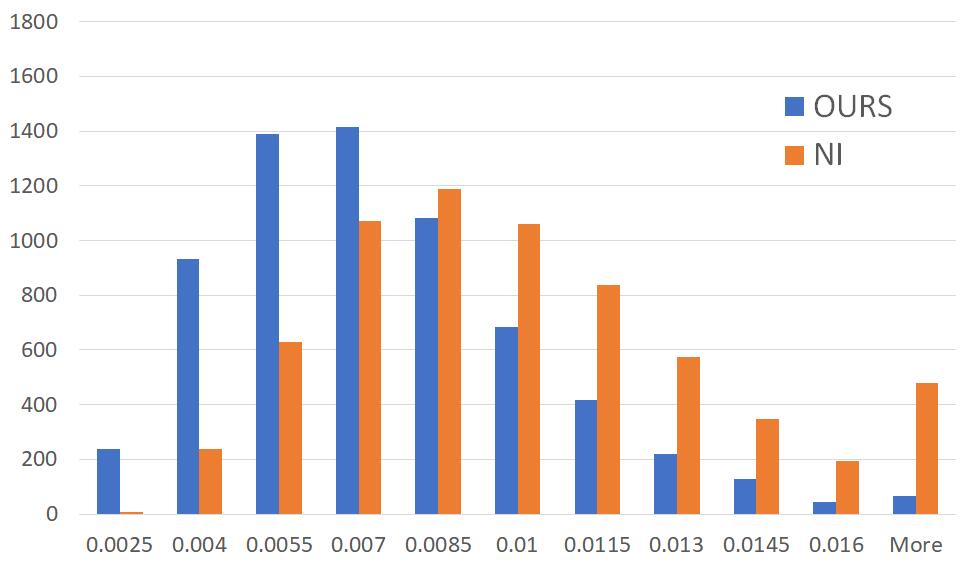}
    \end{minipage}}
	\vspace{-5mm}
	\caption{Distribution histograms of surface and importance errors of on Thingi10K dataset.}
	\label{fig:histograms}
	\vspace{2mm}
    \includegraphics[width=1\linewidth]{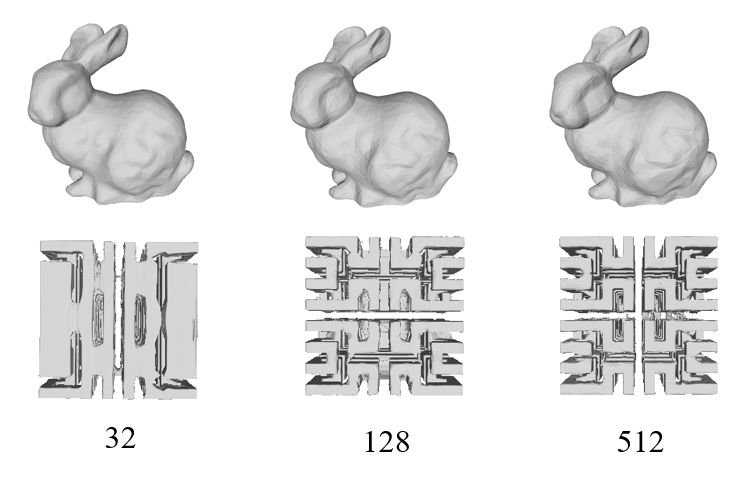}
    \vspace{-10mm}
	\caption{Comparison of our reconstruction results with different numbers of key spheres. From left to right, the spheres number gradually increases. For the simple `bunny' model, all the three reconstructed models  in the first row are visually acceptable. However, the reconstructed model in the second row is not acceptable until using 512 key spheres.}
	\label{fig:different_sphere_number}
\end{minipage}
\end{tabular}
\end{figure}

\vspace{-1mm}
\SubSection{3.1 Influence of Fitting Network Size}\label{4.2}
\vspace{-2mm}
NI has pointed out that reconstruction accuracy can be improved by increasing the number of network parameters.
Thus we compare our method (using DPFE branch) and NI under different fitting network sizes.
Fig.~\ref{fig:layer_hidden} shows the surface errors on the Thingi32 dataset under 16 configurations of the global fitting network. 
No matter what the size of the MLP fitting network is, our method has significantly better reconstruction accuracy than NI.
Fig.~\ref{fig:Statue} also provides a visual comparison for the `Stanford's Lucy' model in Thingi32 under three network configurations. 
Our method can recover better details compared with NI with roughly the same parameters.

We also conduct a comparative experiment on the Thingi10K dataset. Our method and NI adopt $6\!*\!32$ and $8*32$ nodes in global fitting network respectively, which corresponds to 7026 and 7553 parameters. Distribution histograms of surface and importance errors of about 7000 objects are shown in Fig.~\ref{fig:histograms}.
The proposed method shows significantly less errors than NI.

\vspace{-1mm}
\SubSection{3.2 Influence of the Number of Spheres}\label{4.3}
\vspace{-2mm}
In this section, the proposed method with the LPFE branch is used to extract 29D point feature vectors, which guarantees a fair comparison with the NGLOD method utilizing the trained latent vectors of grid points.
As NGLOD proposes several networks with multi-level grid resolution, representing different reconstruct accuracy and storage costs, we also vary the number of key spheres to observe its impact.
The datasets used here are ShapeNet150 and Thingi32, which are evaluated using gIoU and CD. 
The shown numbers of CD are all magnified by 1000 times.
Table~\ref{tab:nglod} shows the average results of our methods with different numbers of key spheres and other methods.
Note that the results of all methods involved in the comparison are from the NGLOD paper.
As shown in Table~\ref{tab:nglod}, we can use much less storage parameters to achieve higher reconstruction accuracy.
Although the LOD5-level NGLOD method shows the highest gIoU in the Thingi32 dataset, its storage capacity for each object ($1356$KB) has greatly exceeded the average storage of Thingi32 ($221$K parameters, which represents $884$KB storage assuming each parameter is stored by $4$ bytes).
Our method can achieve high-fidelity 3D shape compression for all Thingi32 models and most of ShapeNet150 models with only $42$KB storage.
Converting the Thingi32 dataset from mesh format to key spheres based implicit neural representation accomplishes an $\sim$1:21 compression rate, averagely.

Fig.~\ref{fig:different_sphere_number} visualizes the reconstruction results of `bunny' (as a simple model) and `Hilbert cube' (as a complex model) under different numbers of key spheres.
The reconstructed `bunny' (shown on the left of the first row) with the network input of 32 key spheres is visually acceptable, even though using more spheres will improve its fine details.
As shown in the second row, our method fails to properly fit the complex `Hilbert cube' model with 32 or 128 key spheres. When 512 key spheres are used, the details of the reconstructed model become preliminarily acceptable. 

\Section{4. Conclusion}
\vspace{-3mm}
This paper proposes an implicit neural representation of 3D objects based on key spheres which naturally contain the rough signed distance function (SDF) of an object. Therefore, by explicitly inputting multiple key spheres information, our method significantly reduces the difficulty of fitting complex objects through the neural network. 
In addition, the network in our method can directly fit the global model instead of only fitting local shapes in the state-of-the-art method~\cite{nglod}. 
Compared with previous methods, our method can substantially improve the reconstruction accuracy with roughly the same amount of storage parameters.
It is worth noting that the visual attraction of the displayed reconstruction results has obviously exceeded the simplified mesh with the same number of parameters.
Therefore, our method actually achieves
neural compression coding with high fidelity and high compression rate for most of 3D objects in the test dataset.

Future work mainly focuses on two aspects. Firstly, according to the different complexity of objects, we will explore an adaptive way to determine the number of key spheres and the number of network layers. Secondly, we will expand this method to handle large scenarios. Apart from these, other applications based on key spheres representation are also worth investigating. 

\Section{References}
\footnotesize
\bibliographystyle{IEEEbib}
\bibliography{ref_V1_se}

\end{document}